\documentclass[lettersize, journal]{IEEEtran}
\usepackage{amsmath, amsfonts}
\usepackage{algorithmic}
\usepackage{algorithm}
\usepackage{array}
\usepackage[caption=false, font=normalsize, labelfont=sf, textfont=sf]{subfig}
\usepackage{textcomp}
\usepackage{stfloats}
\usepackage{url}
\usepackage{verbatim}
\usepackage{graphicx}
\usepackage{cite}
\newtheorem{myDef}{Definition }
\newtheorem{myTheorem}{Theorem}

\graphicspath{{Pics/}}
\usepackage{hyperref}
\usepackage{booktabs}
\usepackage{multirow}
\usepackage{color}

\usepackage{optidef}  
 
\usepackage{tikz,xcolor,hyperref}
\definecolor{lime}{HTML}{A6CE39}
\DeclareRobustCommand{\orcidicon}{%
    \begin{tikzpicture}
    \draw[lime, fill=lime] (0,0) 
    circle [radius=0.16] 
    node[white] {{\fontfamily{qag}\selectfont \tiny ID}};    \draw[white, fill=white] (-0.0625,0.095) 
    circle [radius=0.007];    \end{tikzpicture}
    \hspace{-2mm}}
\foreach \x in {A, ..., Z}{%
    \expandafter\xdef\csname orcid\x\endcsname{\noexpand\href{https://orcid.org/\csname orcidauthor\x\endcsname}{\noexpand\orcidicon}}
    }

\hypersetup{hidelinks,
	colorlinks=true,
	allcolors=black,
	pdfstartview=Fit,
	breaklinks=true}

\begin{document}


\title{INGB: Informed Nonlinear Granular Ball Oversampling Framework for Noisy Imbalanced Classification}

\author{Min Li\orcidA{}, Hao Zhou\orcidB{}, Qun Liu\orcidC{}, Yabin Shao\orcidD{}, and Guoying Wang\orcidE{},~\IEEEmembership{Senior Member, IEEE}

  \thanks{Min Li, Qun Liu, Yabin Shao, and Guoyin Wang are with the Chongqing Key Laboratory of Computational Intelligence, Chongqing University of Posts and Telecommunications, Chongqing, China. (E-mail: limin\_sky@163.com, liuqun@cqupt.edu.cn, shaoyb@cqupt.edu.cn,
    wanggy@cqupt.edu) (Corresponding author: Qun Liu.)
  }
  \thanks{Hao Zhou is with the Key Laboratory of Dependable Services Computing in Cyber Physical Society-Ministry of Education, College of Computer Science, Chongqing University, Chongqing, China. (E-mail: zhouhaocqupt@163.com) (Co-first author: Hao Zhou.)
  }}

\markboth{IEEE TRANSACTIONS ON NEURAL NETWORKS AND LEARNING SYSTEMS, No.~14, December~2022}%
{Shell \MakeLowercase{\textit{et al.}}: A Sample Article Using IEEEtran.cls for IEEE Journals}

\maketitle

\begin{abstract}
In classification problems, the datasets are usually imbalanced, noisy or complex. Most sampling algorithms only make some improvements to the linear sampling mechanism of the synthetic minority oversampling technique (SMOTE). Nevertheless, linear oversampling has several unavoidable drawbacks. Linear oversampling is susceptible to overfitting, and the synthetic samples lack diversity and rarely account for the original distribution characteristics. An informed nonlinear oversampling framework with the granular ball (INGB) as a new direction of oversampling is proposed in this paper. It uses granular balls to simulate the spatial distribution characteristics of  datasets, and informed entropy is utilized to further optimize the granular-ball space. Then, nonlinear oversampling is performed by following high-dimensional sparsity and the isotropic Gaussian distribution. Furthermore, INGB has good compatibility. Not only can it be combined with most SMOTE-based sampling algorithms to improve their performance, but it can also be easily extended to noisy imbalanced multi-classification problems. The mathematical model and theoretical proof of INGB are given in this work. Extensive experiments demonstrate that INGB outperforms the traditional linear sampling frameworks and algorithms in oversampling on complex datasets.
\end{abstract}


\section{Introduction}
\IEEEPARstart{D}{ata} imbalance has attracted the continuous attention of scholars due to its ubiquity and inevitability. In data imbalance problems, the number of some classes (i.e., the minority class) is smaller or much smaller than that of other classes (i.e., the majority class) \cite{bib1}. This highly skewed data distribution impairs the performance of traditional learning \cite{bib2}. Due to the deficiency of the minority class, learning models are biassed toward the majority class, obtaining a seemingly high accuracy but failing to identify the minority class correctly \cite{bib3}, \cite{bibharm}. In real-world applications, such as disease prediction \cite{bib4}, \cite{bib5}, biometric identification \cite{bib6}, anomaly monitoring \cite{bib7}, \cite{apply2} and diverse visual tasks \cite{bibapply4}, \cite{bibapply5}, the minority class is often more valuable and relevant to the task. Therefore, enhancing minority class recognition without sacrificing the accuracy of the majority class is still a research hotspot.

Generally, class-imbalanced learning methods can be roughly grouped into three categories \cite{bib9}, \cite{bib10}: (1) Cost-sensitive learning assumes a higher misclassification cost for the minority class to emphasize the learning of the minority class \cite{bib11}, \cite{bibcost2}. (2) Algorithmic-level methods modify or ensemble classifiers to efficiently identify minority classes \cite{bibclassifier1}, \cite{bib12}. (3) Preprocessing methods mitigate the effects of skewed class distributions by optimizing the feature space via feature selection \cite{bib13} or by balancing the sample space by resampling \cite{bib14}. Resampling methods are convenient due to their independence from chosen classifiers and specific domains. Resampling methods include oversampling to duplicate or synthesize new minority instances, undersampling to eliminate minority instances, and hybrid sampling to concurrently perform minority instance synthesis and majority instance elimination \cite{bib15}, \cite{bibDeepSMOTE}.

\begin{figure*}
  \centering
  \includegraphics[scale=0.615]{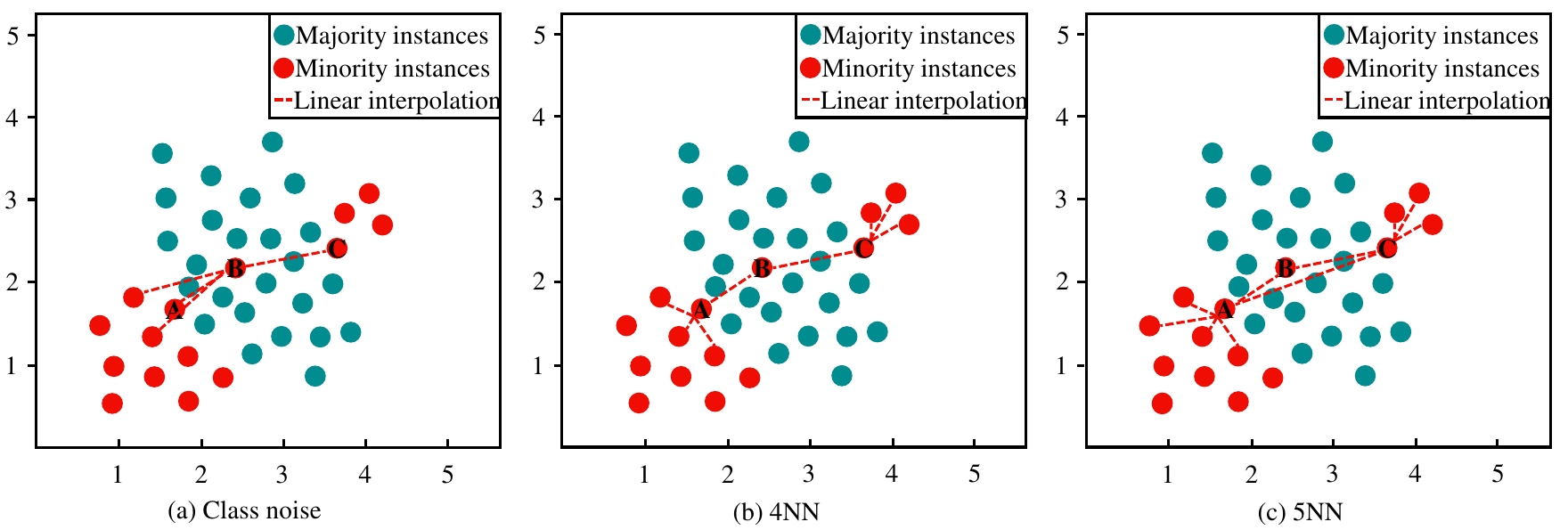}
  \caption{(a) class noise-based linear interpolation with arbitrary $K$, (b) linear interpolation with $K$=4, (c) linear interpolation with $K$=5.}
  \label{intro}
\end{figure*}

The synthetic minority oversampling technique (SMOTE), the most influential resampling algorithm, randomly synthesizes new instances along the lines from a minority instance to its $k$-nearest neighbors \cite{bib16}. Due to its simplicity and effectiveness, SMOTE serves as the established framework for most subsequent oversampling algorithms. Nonetheless, tasks become more difficult when a dataset encompasses other intrinsic characteristics such as class noise, small disjunctions, or class overlapping, particularly for the most disruptive class noise \cite{bib17}, \cite{noise2}. Additionally, the inherent characteristics of SMOTE exacerbate the complexity of datasets with class noise \cite{bib33}. In particular, taking instance B in Fig. \ref{intro}(a) as an example, linearly synthesizing instances with noise as a seed sample introduces additional noisy and overlapping instances, amplifying the performance deterioration. Many SMOTE-based variants, such as area-emphasizing algorithms, noise-filtering algorithms, cluster-based algorithms, and interpolation-based algorithms, have been developed to address these challenges from different perspectives \cite{bibreview}, \cite{bib19}. However, SMOTE and its variants still need to optimize $k$-nearest neighbor hyperparameters. As shown in Fig. \ref{intro}(b) - \ref{intro}(c), an inappropriate $k$-value will introduce more undesirable or noisy instances, which will invade the majority class and worsen the learning complexity. Optimizing the $k$-value has a quadratic time complexity, making it a time-intensive process \cite{bib20}, \cite{bib21}. Furthermore, such a linear interpolation based on $k$-nearest neighbors is susceptible to overlap and lacks the reasonable diversity of synthetic samples. 

To fill these gaps, an efficient and robust nonlinear informed oversampling framework with the granular ball (INGB) for both imbalance and class noise is proposed. First, considering the negative effects of noisy samples and $k$-nearest neighbor hyperparameter optimization, we design an adaptive granular ball generation strategy to obtain quality and robust granular balls that efficiently characterize the data distribution. Specifically, it not only monitors the characteristics of class noise, but also searches the data space completely and effectively. Then, the informative minority balls are selected to perform quality nonlinear sampling by integrating the diverse classes of density-based informed entropy. Finally, following the sparsity and spherical Gaussian distribution, abundant and diversified synthetic samples are generated within each informative granular ball.
The main contributions of this paper are as follows:

\begin{enumerate}[]
  \item The INGB framework is a novel framework that samples and improves classification performance on noisy imbalanced datasets. The mathematical model and theoretical proof of INGB are given in this work.
  \item An adaptive granular ball generation strategy is proposed to characterize the data distribution in noisy imbalanced datasets. It extends the $k$-nearest neighborhood to the high-dimensional granular domain, which contributes to robust and diverse sampling.
  \item A reasonable and diversified nonlinear sampling is conducted within the granular balls by following high-dimensional sparsity and spherical Gaussian distribution. Not only does it prevent spatial over-generalization, but it also enhances diversity while obeying the original distribution.
\end{enumerate}

The rest of this paper is organized as follows. The previous work related to this paper is reviewed in Section \ref{related_work}. The proposed INGB framework is explained in detail in Section \ref{INGB}. Section \ref{experiments} presents extensive experiments and analyses the compare INGB with state-of-the-art resampling algorithms and frameworks. In Section \ref{conclusion}, a brief conclusion is provided.

\section{Related work}\label{related_work}
By far, the most prevalent resampling paradigms are based on the linear sampling mechanism of SMOTE, from which numerous variants have been developed \cite{bib22}. Some studies emphasize specific areas to ensure quality synthetic samples and to prevent the effects caused by generating noise. Safe-level-SMOTE calculates the "security level" of each minority instance by the number of minority instances in its $k$-nearest neighbors and interpolates carefully to maintain more synthetic instances within the safer region \cite{bib23}. Nevertheless, it favors dense minority areas that are susceptible to overlapping and over-generalization. He et al. proposed an adaptive synthetic sampling method (ADASYN), which weights distinct minority instances based on learning difficulty and generates more synthetic instances for the harder-to-learn minority instances \cite{bib24}. Numerous algorithms inspired by ADASYN employ similar mechanisms to regulate the number of synthetic instances for reasonable generalization \cite{bib25}, \cite{bib26}. Barua et al. proposed a majority-weighted minority method (MWMOTE) to identify and weight informative minority instances by the $k$-neighborhood of the minority and majority classes \cite{bib27}. Then, the informative minority instances are selected to synthesize new instances within clusters. However, MWMOTE requires optimization of several $k$ hyperparameters. Adaptive-SMOTE can adaptively divide minority classes into dangerous and inner subsets by the $k$-nearest neighbor interval and oversample the subsets, respectively \cite{bib28}. In addition to being constrained by several hyperparameters, these area-emphasizing algorithms also tend to overlap and overgeneralize in specific areas. Furthermore, the existence of noise might corrupt certain regions or samples, further increasing the complexity of learning.

Noise-filtering algorithms identify and purge noisy or undesirable instances by designing diverse filtering strategies. Batista et al. first advocated combining data cleaning with oversampling to exclude intrusive heterogeneous instances and ensure better-defined classes \cite{bib29}. SMOTE-TomekLinks eliminates multiclass instances that compose Tomek linkages, whereas SMOTE-ENN filters any instance misclassified by its three-nearest neighbors. Then, an iterative-partitioning filter (SMOTE-IPF) identifies and eliminates noisy instances in multiple iterations until the number of identified noises reaches a stopping threshold \cite{bib30}. Recently, a synthetic minority oversampling technique based on Gaussian mixture model filtering (GMF-SMOTE) was proposed to filter out "noisy samples" and "boundary samples" in the grouped subclasses by an expectation-maximum filtering algorithm \cite{bib32}. Although reasonable data filtering is feasible, it is incapable of preventing the additional noise introduced by SMOTE and is more susceptible to being constrained by filtering iterative cleaning. Therefore, Chen proposed self-adaptive robust SMOTE (RSMOTE) to adaptively identify and filter noise by measuring the relative density of its $k$-neighborhoods \cite{bib33}. Then, linear interpolation is performed within noise-free clusters. The classical SMOTE algorithm and most of its variants linearly interpolate synthetic samples in the $k$ neighborhood, in particular, the performance of some algorithms depends on the $k$ neighborhood \cite{bib23}, \cite{bib27}, \cite{bib28}, \cite{bib33}. It not only restricts the diversity of synthetic samples, but also exacerbates their overlap.

Some recent studies suggest enhanced drop-in strategies to overcome the drawbacks of linear interpolation mechanisms. Zhou proposed a general weighting framework (W-SMOTEs) to alleviate the interpolation blindness of SMOTE and its variants \cite{bib34}. It distinguishes noisy samples and carefully determines where to interpolate synthetic samples by calculating the relative chaotic level of each minority instance and its $k$ neighbors. Furthermore, li proposed a weighted space division framework (SW) that adaptively guides the selection of informed neighbors and the assignment of suitable interpolation positions \cite{bib35}. The decision boundary is strengthened, and noise-induced performance deterioration is decreased. Geometric SMOTE (G-SMOTE) extends linear generation mechanism by generating diverse instances inside a safe geometric region selected for each minority instance \cite{bib36}. Nevertheless, selecting a safe geometric region is inseparable from the required parameter turning.

\section{The proposed framework} \label{INGB}
\subsection{Motivation}
\begin{figure*}
  \centering
  \includegraphics[scale=0.4]{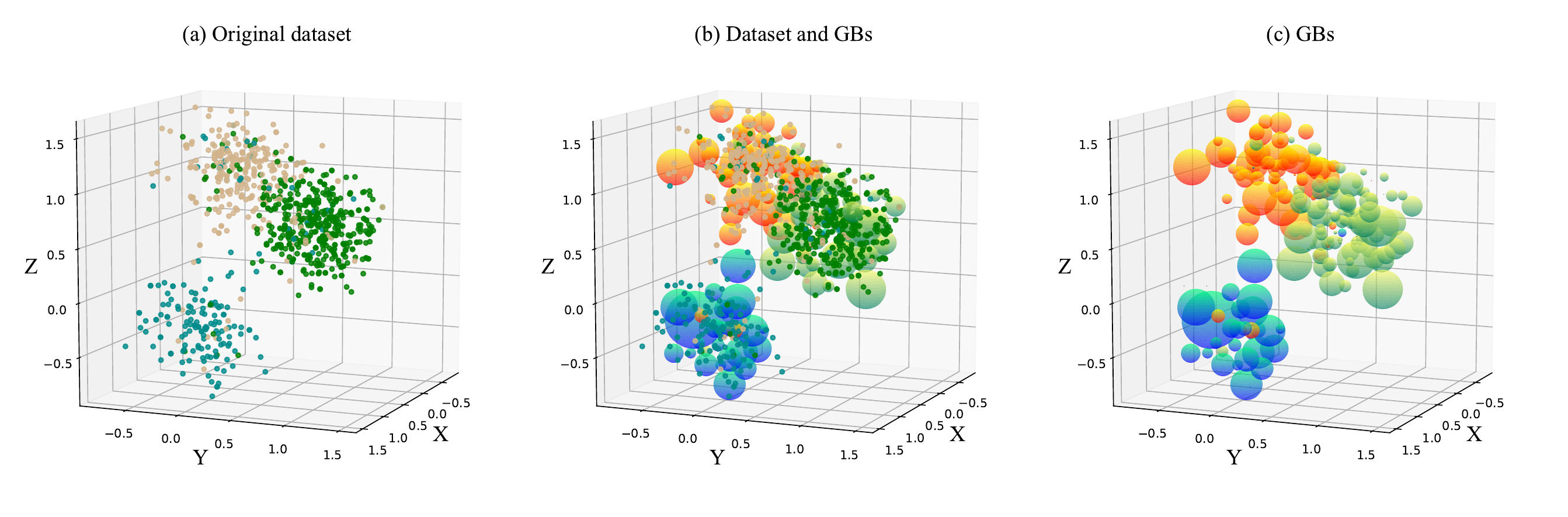}
  \caption{(a) Spatial distribution of 3D triple classification dataset. (b) Hypersphere adaptive simulate dataset's spatial distribution. (c) Spatial distribution after simplification.}
  \label{fig_motivation}
\end{figure*}




Some classical or state-of-the-art sampling algorithms are mostly based on SMOTE. Their ideas are confined to SMOTE and inherit some drawbacks of SMOTE. They also rarely provide mathematical models and deduction processes. There are various complex datasets in reality, such as high-dimensional, large-scale, and highly imbalanced ratio datasets. It is difficult for SMOTE and its variants to identify the distribution characteristics of a dataset by KNN. The linear synthesis of new samples among neighbors tends to cause overfitting. The information gain obtained is usually low. In various practical applications, it is not difficult to find that the performance of SMOTE variants fluctuates seriously. Changes in the datasets can greatly affect their performance. The robustness and generality of linear oversampling need to be improved. Inspired by \cite{bib31}, we propose a new sampling direction, nonlinear sampling, instead of just improving the traditional direction. A data structure that is suitable for complex datasets is given. It can check for noises and outliers while fitting the characteristics of data distribution. And this data structure can summarize the data information in it with a small number of features. In the sample generation phase, new samples have higher information entropy. To prove the reasonability of the framework, we provide the mathematical models and theoretical proofs.

As shown in Fig.\ref{fig_motivation}, Fig.\ref{fig_motivation}(a) is a visualization of a 3D triple classification dataset. This dataset is an imbalanced dataset with noise. The hypersphere in Fig.\ref{fig_motivation}(b) (dimension = 3 in this case) is the data structure we propose. These hyperspheres fit the spatial distribution of the dataset. Fig.\ref{fig_motivation}(c) reduces the dataset to hyperspheres so that noise and outliers can be effectively filtered out. Larger spheres are safer and better for generating new samples. However, smaller spheres are often distributed at the junction of distribution classes or at the edge of classes requiring caution.

\subsection{Granular Ball}
\begin{figure}[htbp]
  \centering
  \includegraphics[scale=0.6]{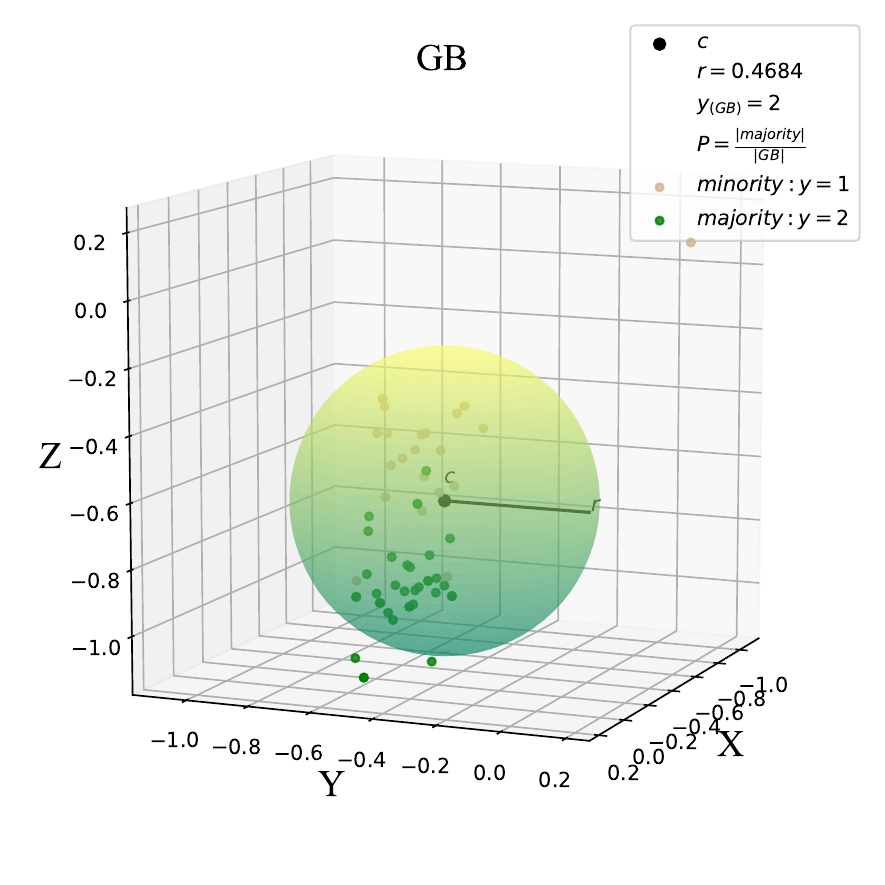}
  \caption{Visualization of $GB$ in 3D space}
  \label{fig_GB}
\end{figure}

\begin{myDef}[Granular Ball]
  \label{def_GB}
  Given an $m*n$ dataset $D = \left\{(x_i, y_i) \;|\; 1 \leq i \leq m \right\}$ , $|D|= m$. Its input space is $X = \left\{ x_i=(x_i^1, x_i^2, x_i^3,..., x_i^n)^T \;|\;1 \leq i \leq m \right\}$, and $x_i \in X \in R^n$. Its output space $Y=\left\{y_i \;|\; 1 \leq i \leq m \right\}$. $D$ is composed of $k$ classes and $D = \left\{D_1, D_2,..., D_k \right\}$.  We introduce the novel data structure of the granular ball, abbreviated as $GB$. It is a hypersphere generated on dataset D. It possesses the following properties: the number of samples $|GB|$, dimension $d$, center $c$, radius $r$, label $y_{(GB)}$, and state $\omega$. Its properties are defined as follows:
  \begin{equation}
    \label{GB}
    GB=\left\{
    \begin{aligned}
      |GB|     & = m, k \; classes;                                                       \\
      d        & = n;                                                                     \\
      x_i, y_i & = \left\{(x_i, y_i) \;|\; 1 \leq i \leq m \right\};                      \\
      c        & = \frac{1}{m} \sum_{i=1}^m x_i ;                                         \\
      r        & = \frac{1}{m} \sum_{i=1}^m (\sum_{l=1}^n{|x_i^l -c^l|^p})^\frac{1}{p};   \\
      y_{(GB)} & = { \underset{x\in [1, k]}{{\arg\max} \, f(x)} = |D_x|. };               \\
      \omega        & = \frac{|D_{y_{(GB)}}|}{m} .                                           &
    \end{aligned}
    \right.
  \end{equation}
\end{myDef}
As shown in Fig.(\ref{fig_GB}), center $c$ is the average of all $X$, and radius $r$ is the average distance of all samples to $c$ (the Euclidean distance is taken as an example hence let $p=2$). The label $y_{(GB)}$ is the label of the majority class in this $GB$. The state $\omega$ is the proportion of majority sample number to the total number of samples.

\begin{myDef}[$\Gamma(x)$]
  \label{Gamma}
  $\Gamma(x) = \int_{0}^{\infty} e^{-t}t^{x-1} dt, $ for $x > 0$. Let $\gamma = \int_{0}^{\infty} e^{-x^2}$, then
  \begin{enumerate}
    \item[(a)] $\Gamma(x) = \int_{0}^{\infty} e^{-x^2} dx = \frac{\sqrt{\pi}}{2}$;
    \item[(b)] $\Gamma(\frac{1}{2}) = 2\gamma = \sqrt{\pi}$;
    \item[(c)] $\Gamma(x+1) = x\Gamma(x), for \ x > 0, \Gamma(n) = (n-1)! \ if n \in N$;
    \item[(d)] The volume of a $d$ dimensional unit sphere is $\frac{\pi^{d/2}}{\Gamma(\frac{d}{2}+1)}$
  \end{enumerate}
\end{myDef}

\begin{myTheorem}[$V_n$]
  \label{Theorem of V_n}
  Given a hypersphere of dimension $n$ with radius $r$, its volume $V_n$ is expressed as follows:
  \begin{equation}
    \label{V_n}
    V_n = \frac{{{\pi}^{\frac{n}{2}}}r^{n}}{\Gamma(\frac{n}{2} + 1)}
  \end{equation}
\end{myTheorem}

\begin{IEEEproof}
  \label{proof of Vn}
  \newline
  For $n = 2$, the volumn is computed by Eq.\eqref{V2}
  \begin{equation}
    \label{V2}
    V_2 = \int_{0}^{r} \int_{0}^{2\pi}r drd\theta = \pi R^2
  \end{equation}
  For $n=3$, the volumn is computed by Eq.\eqref{V3}
  \begin{equation}
    \label{V3}
    V_3 = \int_{0}^{r}  \int_{\frac{-\pi}{2}}^{\frac{\pi}{2}} \int_{0}^{2\pi}r^2 cos\theta_1 drd\theta_1d\theta_2 = \frac{4\pi r^3}{3}
  \end{equation}
  For $n \geq 4$, the volumn is computed by
  \small{
    \begin{eqnarray}    \label{eq}
      V_n \! &=& \! \int_{0}^{r}\! \int_{0}^{2\pi}\! \int_{\frac{-\pi}{2}}^{\frac{\pi}{2}}\! {\cdots} \! \int_{\frac{-\pi}{2}}^{\frac{\pi}{2}} \!  [r^{n-1}cos^{n-2}\theta_1 cos^{n-3}\theta_2 {\cdots}] \! d\theta_1d\theta_2 {\cdots} dr  \nonumber \\
      ~&=&\frac{{{\pi}^{\frac{n}{2}}}r^{n}}{\Gamma(\frac{n}{2} + 1)}
    \end{eqnarray}

  }

  Note that the above computations exploit the properties of $\Gamma(x)$ defined in Definition \ref{Gamma} and follow Beta functions and trigonometry:

  \begin{eqnarray}
    Beta(\alpha,\beta) = \int_{0}^{1} x^{\alpha-1}(1-x)^{\beta-1}dx, \alpha,\beta > 0  \nonumber    \\
  \end{eqnarray}
  obviously, we can obtain
  \begin{eqnarray}
    Beta(\alpha,\beta)=\frac{\Gamma(\alpha)\Gamma(\beta)}{\Gamma(\alpha + \beta)}
  \end{eqnarray}

  \begin{eqnarray}
    \int_{\frac{-\pi}{2}}^{\frac{\pi}{2}} cos^{m}\theta d\theta &=& \int_{0}^{\frac{\pi}{2}} 2cos^{m} \theta d\theta  \nonumber    \\
    ~&=& \int_{0}^{1} 2x^{m} \frac{1}{-\sqrt{1-x^2}}dx, where \;x = cos\theta \nonumber    \\
    ~&=& \int_{0}^{1} 2y^{\frac{m}{2}}(1-y)^{\frac{-1}{2}} \frac{1}{2\sqrt{y}} dy, where \;x = \sqrt{y} \nonumber    \\
    ~&=& \int_{0}^{1} 2y^{\frac{m+1}{2}-1} (1-y)^{\frac{1}{2}-1}dy \nonumber \\
    ~&=& Beta(\frac{m+1}{2},\frac{1}{2}) \nonumber \\
    ~&=& \frac{\Gamma(\frac{m+1}{2})\Gamma(\frac{1}{2})}{ \Gamma(\frac{m}{2}+1)}
  \end{eqnarray}

  Therefore,
  \begin{equation}
    V_n = \frac{{{\pi}^{\frac{n}{2}}}r^{n}}{\Gamma(\frac{n}{2} + 1)}
  \end{equation}

\end{IEEEproof}

\subsection{Splitting Granular Balls}
\subsubsection{Method for $GB$ Splitting}
\begin{myDef}[Necessary and insufficient conditions for splitting]
  \label{def_split}
  \begin{enumerate}
    \item[(a)] $|GB| \geq k(n+1)$. $GB$ will be splited into $k$ $GB(sub)$ and $|GB(sub)|$ $\geq (n+1)$. Otherwise an n-dimensional hypersphere can not be constructed, as stated in Theorem \eqref{Theorem of n_dimensional}. Therefore, $|GB|$ must have at least $k(n+1)$ samples.
    \item[(b)] State $\omega \geq T$. The current $GB's$ $\omega$ is less than the lowest state bound $T$.
  \end{enumerate}
\end{myDef}

When we obtain a dataset, we first consider the whole dataset as an initial $GB$ and calculate each property, referring to Definition \ref{def_GB}. The state lower bound $T$ is a pre-given parameter. $|GB|$ and $\omega$ are two crucial parameters to determine whether splitting occurs. $GB$ must satisfy the above two necessary conditions for the granular ball to be split. If a $GB$ has $k$ classes, it will be split into $k$ $GB(sub)$ with the same $n$-dimensions. Then, it will be split iteratively until all the $GB$ can no longer be split.

\subsubsection{Mathematical Principle of $GB$ splitting}

\begin{myTheorem}
  \label{Theorem of n_dimensional}
  Determining an $n$ dimensional space requires ($n+1$) points, and these $n+1$ points do not share $n-1$ space. Similarly, for an $n$-dimensional $GB$, $|GB| \geq n+1$.
\end{myTheorem}

\begin{IEEEproof}

  There exist $n+1$ points, one of which is chosen as the coordinate origin and the other $n$ points are represented by vectors. These $n$ vectors, each with $n$ components, form an $n*n$ square matrix $A$.
  \begin{equation}
    \label{eqa_Ax=b}
    Ax = b
  \end{equation}

  If $R(A) = n$, $A$ is invertible. The system in Eq. \eqref{eqa_Ax=b} has only one solution. An $n$-dimensional space can be determined. Any vector can be expanded with these $n$ vectors.

  If $R(A) < n$, the system in Eq. \eqref{eqa_Ax=b} has an infinite number of solutions. It is impossible to determine an $n$-dimensional space. The points in the subspace are mapped as zero vectors. In the case of a $3$ dimensional space, if $4$ points are selected to be coplanar, there is a normal vector whose projection on this plane is $0$.

  Hence, to construct an $n$-dimensional $GB$, at least $n + 1$ samples are needed.
\end{IEEEproof}

Suppose a $GB$ is generated according to Definition \ref{def_GB} and satisfies the splitting condition of Definition \ref{def_split}. Samples in a $GB$ are split into $k$ $GB(sub)$ and each sample belongs to only one $GB(sub)$ (no overlap of $GB(sub)$). Our goal is to group similar points into one $GB(sub)$. In this work, we use the distance metric to determine the similarity with the parameter $p=2$ as an example. In this case, the Euclidean distance is chosen. Therefore the splitting of $k$ disjoint non-empty sets is obtained as follows:
$$\left\{(x_i, y_i) \;|\; 1 \leq i \leq m \right\} = \left\{GB(sub_j) \;|\; 1 \leq j \leq k \right\}$$
The sample distances in the same $GB(sub)$ should be as small as possible. We define the sum of squares of the intra-sphere distances as:
$$ W(GB(sub_1), GB(sub_2) ,..., GB(sub_k)) = \sum_{j=1}^{k}\sum_{x \in GB(sub_j)}^{} \Vert x - c_j\Vert^{2}$$
where $c_i$ is the center of the $i$th $GB(sub)$. Based on the above descriptions, we propose an optimization model for $GB$ splitting in Definition \ref{def_gb_split}.

\begin{myDef}[optimization model]
  \label{def_gb_split}
  Given a $GB$ that satisfies the splitting conditions is generated according to Definition \ref{def_GB}. We need to find a splitting strategy so that the sum of the squared distances within the $GB(sub)$ is minimized. Each $|GB(sub)| \geq n+1$, $\omega \geq T$. There is no overlap among all $GB(sub)$. The optimization model is as follows:
  \begin{mini*}|s|
    {GB(sub_j)}{\sum_{j=1}^{k}\sum_{x \in GB(sub_j)}^{} \Vert x - c_j\Vert^{2}}
    {}{}
    \addConstraint{|GB(sub_j)| \geq n+1}
    \addConstraint{\omega_j \geq T}
    \addConstraint{\left\{(x_i, y_i) \;|\; 1 \leq i \leq m \right\} = \left\{GB(sub_j) \;|\; 1 \leq j \leq k \right\}}
    \addConstraint{GB(sub_i) \cap GB(sub_j) = \emptyset, \; \forall i \neq j}{}
  \end{mini*}
\end{myDef}
This is a combinatorial optimization problem, which is an NP-hard problem and is solved by an iterative approach.

\subsection{Oversampling in Granular Balls}
\subsubsection{Method for $GB$ Oversampling}
After the iterative splitting of granular balls, a series of $GB$ are available, which fully reflect the data distribution information. The state boundary $T$ needs to be adjusted to obtain granular balls that meet the requirements, but the state threshold does not restrict our framework. Instead, it is set to 1 to ensure high-quality $GB$ which adapts to the distributions of various datasets. In addition, to further ensure the synthesis of informative samples, information content carried by the minority granular balls is measured by the weighted information entropy. In information theory, entropy measures the expected average amount of information contained within a set \cite{bib37}. Therefore, it is frequently employed as a measure of information content \cite{bib1, bib38}. The instances in a $GB$ with greater entropy contain more "information" than those in a $GB$ with lower entropy. That is, the $GB$ is more uncertain, and vice versa. Therefore, entropy is a good representation of the $GB's$ information content.

In binary datasets, the $GB$, where $y_{GB}$ equals to the minority, is selected. In multiclass datasets, each round selects the $GB$ to be interpolated based on the label that need to be sampled. Suppose the dataset $D$ is composed of the majority class subset $D_{maj}$ and  the minority class $D_{min}$. There exists a set $I$ consisting of minority $GB$, $|I| = N_{g}$ , where $y_{(GB_j)}$ is a label to resample.
\begin{equation}
I=\left\{GB_j \;|\; 1 \leq j \leq N_g \right\}.
\end{equation}

\begin{myDef}[Instance-informed Statistic]
  \label{def of IIS}
  For each instance $x_i \in GB_j(1 \leq j \leq N_g)$, the density-based instance-informed statistic of $x_i$ is denoted by $\varphi(x_i)$. It is the ratio of the reciprocal of the average distance between $x_i$ and its intra-class and inter-class instances within $GB_j$. $\varphi(x_i)$ is defined as follows:

  \begin{equation}
    \resizebox{0.89\hsize}{!}{$
        \varphi(x_i) =\left\{
        \begin{aligned}	
           & \frac{|homo(x_i)|}{\sum_{}dist(x_i,homo(x_i))}, \; |hete(x_i)| = 0,                                                 \\ \\
           & \frac{|homo(x_i)| /{\sum_{}dist(x_i,homo(x_i))}}{|hete(x_i)| / {\sum_{}dist(x_i,hete(x_i))}}, \; |hete(x_i)| \neq 0
        \end{aligned}
        \right.
      $}
  \end{equation}
\end{myDef}
where $homo(x_i)$ and $hete(x_i)$ are the intra- and inter-class information within $GB_j$ of $x_i$, respectively. Note that the class density of $x_i$ is described by the reciprocal of the average distance from $x_i$ to certain classes in the ball, which can effectively reflect the class distribution of $x_i$. A larger ratio results in denser homogeneous versus sparser heterogeneous data distributions around $x_i$.

\begin{myDef}[Ball-Informed Statistic]
  \label{equ of BIS}
  For each $GB$, the entropy-based ball-informed statistic of $GB_j(1 \leq j \leq N_g )$ is denoted by $\delta_j$. It is the expected average amount of density-based information. $\delta_j$ is defined as follows:
  \begin{equation}
    \resizebox{0.89\hsize}{!}{$
        \delta_j = - \frac{1}{N_j}\sum_{i=1}^{N_j} \rho(x_i)\log_2\rho(x_i) \;\; \mbox{s.t.} \;\; \rho(x_i) = \frac{\varphi(x_i)}{\sum_{i=1}^{N_j} \varphi(x_i)}
      $}
  \end{equation}
\end{myDef}
where $N_j$ is the number of instances in $GB_j$, and $\rho(x_i)$ is a percentage of the overall density metric for $x_i$ in $GB_j$, which can be viewed as the probability of $x_i$ in $GB_j$. It can be known that for $GB_j$, the lower the entropy is, the less uncertainty there is, and the more density-based information content carried. Note that $\delta_j$ measures the expected intra-class and inter-class density-based information content.

\begin{myDef}[Seed $GB$] For the set of minority $GB$ $I$, to maximize the availability of adequate and informative synthetic samples, informative sampling $GB$ is determined as seed. There exists a set $S$ consisting of seeds. $S$ is defined as follows:
\begin{equation}
    S = \left\{ GB_j \;|\; 1 \leq j \leq N_g \; , \; \delta_j \geq \frac{1}{N_g}\sum_{j=1}^{N_g} \delta_j \right\}
\end{equation}

\end{myDef}

\begin{myDef}[Sparsity]
  \label{def of Sparsity}
  Given a $GB$ with $n$ dimensions. The volume is calculated by Theorem \eqref{Theorem of V_n} and is denoted as $Vn$. The sparsity of the $GB$ is defined in Eq. \eqref{equ of Sparsity}. Referring to Eq. \eqref{equ of num}, the number of $GB$ to oversample is calculated by the sparsity.
  \begin{equation}
    \label{equ of Sparsity}
    S_{GB} = \frac{|GB|^n}{V_n}
  \end{equation}
  \begin{equation}
    \label{equ of num}
    N_{GB} = \frac{(|D_{maj}| - |D_{min}|)S_{GB}}{\sum_{(GB \in S)}^{} S_{GB}}
  \end{equation}
\end{myDef}

For each sample $x_i$ in the seed $GB$ of $S$, the sample $x_j$ in the same $GB$ is selected randomly as pair. Then the weighted average center and radius of this pair are calculated. Finally oversampling in a isotropic high-dimensional Gaussian distribution is executed.

\subsubsection{Mathematical Principle of $GB$ Oversampling}
The one-dimensional Gaussian distribution is shown in Eq.\eqref{1d gaussian}:
\begin{equation}
  \label{1d gaussian}
  f\left(x \mid \mu, \sigma^{2}\right)=\frac{1}{\sqrt{2 \pi \sigma^{2}}} e^{-\frac{(x-\mu)^{2}}{2 \sigma^{2}}}
\end{equation}

The concept of the Gaussian distribution can be extended to more than one dimension. The probability density function of a general multivariate Gaussian distribution in $n$ dimensions is as follows:
\begin{equation}
  \label{nd gaussian}
  f\left(x_{1}, \ldots, x_{k}\right)=\frac{\exp \left(-\frac{1}{2}(\mathbf{x}-\mu)^{\mathrm{T}} \Sigma^{-1}(\mathbf{x}-\mu)\right)}{\sqrt{(2 \pi)^{k}|\Sigma|}}
\end{equation}
where $|\Sigma|$ is the determinant of the covariance matrix. For 2 dimensions, the mean vector µ and the symmetric covariance matrix $\Sigma$ are defined as follows:
\begin{equation}
  \mu=\left(\begin{array}{l}
    \mu_{1} \\
    \mu_{2}
  \end{array}\right), \quad \Sigma=\left(\begin{array}{cc}
    \sigma_{1}^{2}             & \rho \sigma_{1} \sigma_{2} \\
    \rho \sigma_{1} \sigma_{2} & \sigma_{2}^{2}
  \end{array}\right)
\end{equation}

An isotropic Gaussian distribution (spherical Gaussian distribution) refers to a multidimensional Gaussian distribution with the same variance in all directions, where the covariance is a positive real number multiplied by the identity matrix. Because of the circular symmetry of Gaussian distribution, isotropy is obtained by simply making the lengths of each axis the same, which means that the distribution density value is related only to the point-to-mean distance, not to the direction. Each dimension of the isotropic Gaussian distribution is also independent of each other, so the density equation can be written in the form of a product of several 1-dimensional Gaussian distribution. The number of parameters of this type of Gaussian distribution increases linearly with the dimensions. Only the mean value increases, while the variance is a scalar quantity. The computational and storage requirements are not significant.

We use a spherical Gaussian distribution for oversampling for three reasons:
\begin{enumerate}
  \item[(a)] Many machine learning models are based on the assumption that the data obey a Gaussian distribution (not a strict Gaussian distribution). $GB$ is also a high-dimensional hypersphere.
  \item[(b)] According to the central limit theorem \eqref{Central limit} in probability theory, the limit of many distributions is a Gaussian distribution when the sample size is infinite.
  \item[(c)] In terms of entropy, in the case where the mean and variance of the data are known (the original data distribution type is unknown), the entropy of a Gaussian distribution is the largest among all other distributions. The "maximum entropy" is approximately equivalent to the "closest uniform distribution under the same constraints", i.e., more realistic.
\end{enumerate}

\begin{myTheorem}
  \label{Central limit}
  Let the random variables $x_1$, $x_2$, $x_3$,..., $x_n$ be independent of each other and satisfy the same distribution. $E(X_k) = \mu $ and $D(X_k)=\sigma,(k=1, 2, 3,..., n)$. The random variable $\mathrm{Y}_{n}=\frac{\sum_{k=1}^{n} \mathrm{X}_{k}-n \mu}{\sigma \sqrt{n}}$. Then, when $n$ tends to infinity, the random variable $Y_n$ accord with a standard Gaussian distribution. Denote the distribution function of $Y_n$ by $F_n$. The theorem can be expressed by the following equations:
  \begin{equation}
    \mathrm{Y}_{n}=\frac{\sum_{k=1}^{n} \mathrm{X}_{k}-n \mu}{\sigma \sqrt{n}}
  \end{equation}
  \begin{equation}
    \lim _{n \rightarrow \infty} F_{n}(x)=\int_{-\infty}^{x} \frac{1}{\sqrt{2 \pi}} e^{-t^{2} / 2} d t
  \end{equation}
\end{myTheorem}

\begin{IEEEproof}

  Let $(X, Y)$ be a two-dimensional continuous random variable, and $X, Y$ are independent of each other and their distribution functions are:
  \begin{equation}
    F_{x}(u)=\int_{-\infty}^{u} g(t) d t,\;\; F_{y}(u)=\int_{-\infty}^{u} h(t) d t
  \end{equation}
  If the random variable $Z=X+Y$, then the distribution function of $Z$  is defined as following: 
  \begin{equation}
  F_z(u)=P\left\{ Z<=u \right\}=P \left\{ X+Y<=u \right\}.
  \end{equation}
  \begin{eqnarray}
      F_{z}(u) &=&
      \iint_{x+y \leq u} f(x, y) d x d y \nonumber\\
      &=&\int_{-\infty}^{+\infty}\left[\int_{-\infty}^{u-y} f(x, y) d x\right] d y  \nonumber\\
      &=&\int_{-\infty}^{+\infty}\left[\int_{-\infty}^{u} f(t-y, y) d t\right] d y \nonumber\\
      &=&\int_{-\infty}^{u}\left[\int_{-\infty}^{+\infty} f(t-y, y) d y\right] d t \nonumber\\
      &=&\int_{-\infty}^{u}\left[\int_{-\infty}^{+\infty} g(t-y) h(y) d y\right] d t
  \end{eqnarray}
  Then the probability density of $Z$, $f_{z}(u)$, is:
  \begin{equation}
    \label{fzu}
    f_{z}(u)=\int_{-\infty}^{+\infty} g(t-y) h(y) d y \stackrel{F}{\longleftrightarrow} G(j w) H(j w)
  \end{equation}

Two conclusions can be got based on Eq. \eqref{fzu}:
  \begin{enumerate}
    \item [(a)] The probability density function of the sum of two independent and continuous random variables is the convolution of their respective probability density functions.
    \item [(b)] From the convolution theorem, we know that the Fourier transform of the probability density function of the sum is equal to the product of the Fourier transforms of the respective probability density functions.
  \end{enumerate}
  Let $a, b$ be real numbers, then the distribution function of $\frac{(X+b)}{a}$ is:
  \begin{equation}
    F_{\frac{x+b}{a}}(u)=P\left\{\frac{X+b}{a} \leq u\right\}=a \int_{-\infty}^{u} g(a t-b) d t
  \end{equation}
  The probability density and the corresponding Fourier transform are:
  \begin{eqnarray}
    f_{\frac{x+b}{a}}(u) &=&
    a g(a u-b) \stackrel{F}{\longleftrightarrow} \int_{-\infty}^{+\infty} a g(a u-b) e^{-j w u} d u \nonumber\\
    &=&\int_{-\infty}^{+\infty} g(t-b) e^{-j \frac{w}{a} t} d t \nonumber\\
    &=&\int_{-\infty}^{+\infty} g(t) e^{-j \frac{w}{a}(t+b)} d t \nonumber\\
    &=&e^{-j w \frac{b}{a}} G\left(j \frac{w}{a}\right)
  \end{eqnarray}
  Then according to the condition of the central limit theorem, combined with the previous conclusions, the probability density of $Y_n$ and its Fourier transform are obtained:
  \small{
    \begin{eqnarray}
      \label{Yn}
      f_{Y n}(u) \stackrel{F}{\longleftrightarrow} Y_{n}(j w) & = & \left(e^{-j w \frac{-\mu}{\sigma \sqrt{n}}} G\left(j \frac{w}{\sigma \sqrt{n}}\right)\right)^{n} \nonumber\\
      & = & \left(\int_{-\infty}^{+\infty} g(t) e^{-j \frac{w}{\sigma \sqrt{n}}(t-\mu)} d t\right)^{n}
    \end{eqnarray}
  }
  The Taylor decomposition of the product function in the expansion of Eq.\eqref{Yn} is written below:
  \small{
    \begin{gather}
      \int_{-\infty}^{+\infty} g(t) e^{-j \frac{w}{\sigma \sqrt{n}}(t-\mu)} d t \nonumber\\
      =\int_{-\infty}^{+\infty} g(t)\left[1+\left(-j \frac{w}{\sigma \sqrt{n}}(t-\mu)\right)+\frac{\left(-j \frac{w}{\sigma \sqrt{n}}(t-\mu)\right)^{2}}{2}\right] d t \nonumber\\
      =1-\frac{w^{2}}{2 n}
    \end{gather}
  }
  When $n$ tends to infinity, the probability density of $Y_n$ can be found according to the Fourier inverse transform.
  \begin{eqnarray}
    \lim _{n \rightarrow \infty} f_{Y n}(u)&=&\frac{1}{2 \pi} \int_{-\infty}^{+\infty}\left[\lim _{n \rightarrow \infty} Y_{n}(j w)\right] e^{j w u} d w \nonumber\\
    ~&=&\frac{1}{2 \pi} \int_{-\infty}^{+\infty}\left[\lim _{n \rightarrow \infty}\left(1-\frac{w^{2}}{2 n}\right)^{n}\right] e^{j w u} d w \nonumber\\
    ~&=&\frac{1}{2 \pi} \int_{-\infty}^{+\infty} e^{-\frac{w^{2}}{2}} e^{j w u} d w \nonumber\\
    ~&=&\frac{1}{2 \pi} e^{-\frac{u^{2}}{2}} \int_{-\infty}^{+\infty} e^{\frac{(j w+u)^{2}}{2}} d w \nonumber\\
    ~&=& \frac{1}{\sqrt{2 \pi}} e^{-\frac{u^{2}}{2}}
  \end{eqnarray}
  Hence, the final distribution of $Y_n$ is obtained.
  \begin{equation}
    \lim _{n \rightarrow \infty} F_{n}(x)=\int_{-\infty}^{x} \frac{1}{\sqrt{2 \pi}} e^{-t^{2} / 2} d t
  \end{equation}
\end{IEEEproof}

\subsection{Time Complexity Analysis}
The time complexity of the INGB framework is determined by two main parts: granular ball splitting and granular ball oversampling. Given a dataset $D$ containing $N$ samples and $n$ minority samples ($n \textless \frac{1}{k} N $) with $k$ classes is divided into $N_g$ $GB$. For each iteration of $GB$ splitting, assume that $GB_1, GB_2,..., GB_j$ have the number of samples $N_1, N_2,..., N_j$. The division requires $kN_1+kN_2+...+kN_j \leq kN$ computations. Therefore, the time complexity of granular ball splitting is $O(kNt)$, where $t$ is the number of iterations. In the worst case, even with just one sample per $GB$, $\log_kN$ iterations are needed. However, the imbalanced data distribution makes it impossible to thoroughly mix diverse data classes. Consequently, the number of iterations $t$ is often well below the dataset magnitude and can be viewed as a constant. Therefore, the time complexity of granular ball splitting is linear. For granular ball oversampling, assume that $GB_1, GB_2,..., GB_s$ ($s \textless N_g$) have the number of total samples $N_1, N_2,..., N_s$ and minority samples $n_1, n_2,..., n_s$, respectively. Entropy-based instance-informed and ball-informed statistics require that $n_1*N_1+n_2*N_2+...+n_s*N_s \leq 2n_1^{2}+2n_2^{2}+...+2n_s^{2} \leq 2n^{2}$, since $N_i \textless \frac{1}{2} n_i$. The time complexity of sampling within seed $GB$ by an isotropic high-dimensional Gaussian distribution is $O(nl)$, where $l$ is the number of each $GB$ to oversample. Finally, the overall time complexity of the INGB framework is $O(kN+n^{2})$. The specific pseudocode is shown in Algorithm \ref{The INGB framework}.

\begin{algorithm}[H]
  \caption{The INGB framework.}
  \label{The INGB framework}
  \begin{algorithmic}[1]
    \REQUIRE Imbalanced data set $D$, state lower bound $T$. \\
    \ENSURE Balanced data set $D'$.\\

    //\textbf{First Step} : Granular Ball Splitting.
    \STATE Initialize the $GB$ with the entire data set $D$;
    \STATE Use a Vector V to hold all $GB$, $V = \left\{ GB \right\}$;
    \STATE Initialize the index, int i = 0;
    \WHILE {$i \leq V.size()$}
    \STATE $V$.pop($GB_i$);
    \STATE Compute all properties of $GB_i$ according to Def. \eqref{def_GB};\\
    \IF  {The conditions of Def. \eqref{def_split} are satisfied}
    \STATE Divide $GB_i$ into $k$ $GB(sub)$ of the same dimension;
    \STATE $GB_i = \left\{GB(sub_j) \;|\; 1 \leq j \leq k \right\}$;
    \STATE $V$.push($\left\{GB(sub_j) \;|\; 1 \leq j \leq k  \right\}$);
    \ELSE
    \STATE i++;
    \ENDIF
    \ENDWHILE

    //\textbf{Second Step:} Oversampling in $GB$;
    \STATE Initialize $D_{new}$ = $\emptyset$;
    \STATE Filter out the $GB$ to be oversampled based on the labels;
    \STATE $I=\left\{GB_j \;|\; 1 \leq j \leq N_g, \;, \; y_{(GB_j)} \; is \; a \; label \; to \; resample \right\}$;
    \STATE Compute instance-informed statistic $\varphi(x_i)$ of all granular balls in $I$ by Def.\eqref{def of IIS};
    \STATE Compute ball-informed statistic $\delta_j$ with informed entropy by Def.\eqref{equ of BIS};
    \STATE  seed $GB$: $ S = \left\{ GB_j \;|\; 1 \leq j \leq N_g \; and \; \delta_j \geq \frac{1}{N_g}\sum_{j=1}^{N_g} \delta_j \right\}$.
    \STATE Compute $S_{GB}$ of $GB$ in $S$ by Def.\eqref{def of Sparsity} and Eq.\eqref{equ of Sparsity};
    \STATE Compute $N_{GB}$ of $GB$ in $S$ by Eq.\eqref{equ of num};

    \FOR {each $GB \in S$}
    \WHILE{$N_{GB} != 0$}
    \FOR {each sample $x_i \in GB$}
    \STATE randomly select $x_j$ in the same $GB$ as pair;
    \STATE Compute the pair's weighted average center and radius;
    \STATE Synthesize a new sample $S_{new}$ in a isotropic high-dimensional Gaussian distribution;
    \STATE $D_{new} = D_{new} \cup S_{new}$ ;
    \STATE $N_{GB}$-\;-;
    \ENDFOR
    \ENDWHILE
    \ENDFOR
    \STATE $D' = D \cup D_{new} $;

    \STATE \textbf{Return} $D'$ .
  \end{algorithmic}
\end{algorithm}

\section{Experiments} \label{experiments}

\subsection{Experimental Settings}
To empirically evaluate the effectiveness and robustness of the proposed INGB framework, a series of experiments are conducted to answer the following research questions:
\begin{itemize}
  \item Q1: How does the INGB framework improve the performance of representative resampling compared to state-of-the-art resampling frameworks?
  \item Q2: How does the INGB framework compare to mainstream resampling algorithms?
  \item Q3: Is the INGB framework statistically superior to state-of-the-art resampling algorithms and frameworks?
  \item Q4: How scalable is the INGB framework on multiclass datasets?
\end{itemize}

\subsubsection{Datasets} To demonstrate the general applicability of the INGB framework, 29 real-world benchmark datasets available from the UCI repository \cite{bib39} and KEEL-dataset repository \cite{bib40} are utilized. Table \ref{datasets} provides the basic information of 22 binary datasets with distinct data sizes and dimensions and the imbalance ratio (IR) ranging from 3.25 to 104.81. Table \ref{datasets_multi} provides the basic information of seven multiclass datasets with distinct data sizes and dimensions and different class distributions. To demonstrate the robustness of the INGB framework, instances with $nr$ percent of both the minority and majority classes are randomly flipped to heterogeneous instances (0\% $\leq \ nr \leq $ 30\%).

\begin{table}[htbp]
  \centering
  \caption{Information of the binary datasets}
  \label{datasets}
  \scalebox{0.86}{
    \begin{tabular}{lccccr}
      \toprule
      Datasets                       &
      Samples                        &
      Features                       &
      Majority                       &
      Minority                       &
      IR
      \\
      \midrule
      new-thyroid1                   &
      215                            &
      5                              &
      180                            &
      35                             &
      5.14
      \\
      ecoli                          &
      336                            &
      7                              &
      284                            &
      52                             &
      5.46
      \\
      wisconsin                      &
      488                            &
      9                              &
      444                            &
      44                             &
      10.09
      \\
      diabetes                       &
      550                            &
      8                              &
      500                            &
      50                             &
      10.00
      \\
      breastcancer                   &
      532                            &
      10                             &
      444                            &
      88                             &
      5.05
      \\
      vehicle2                       &
      690                            &
      18                             &
      628                            &
      62                             &
      10.13
      \\
      vehicle                        &
      846                            &
      18                             &
      647                            &
      199                            &
      3.25
      \\
      yeast1                         &
      1160                           &
      8                              &
      1055                           &
      105                            &
      10.05
      \\
      Faults                         &
      1392                           &
      33                             &
      1266                           &
      126                            &
      10.05
      \\
      segment                        &
      2310                           &
      16                             &
      1980                           &
      330                            &
      6.00
      \\
      satimage                       &
      3188                           &
      36                             &
      2893                           &
      295                            &
      9.81
      \\
      wilt                           &
      4338                           &
      5                              &
      4265                           &
      73                             &
      58.42
      \\
      svmguide1                      &
      4400                           &
      4                              &
      4000                           &
      400                            &
      10.00
      \\
      mushrooms                      &
      4628                           &
      112                            &
      4208                           &
      420                            &
      10.02
      \\
      page-blocks0                   &
      5472                           &
      10                             &
      4913                           &
      559                            &
      8.79
      \\
      Data\_for\_UCI\_named          &
      7016                           &
      13                             &
      6379                           &
      637                            &
      10.01
      \\
      letter                         &
      11057                          &
      16                             &
      10052                          &
      1005                           &
      10.00
      \\
      avila                          &
      11695                          &
      10                             &
      10632                          &
      1063                           &
      10.00
      \\
      magic                          &
      13565                          &
      10                             &
      12332                          &
      1233                           &
      10.00
      \\
      susy                           &
      23752                          &
      18                             &
      21593                          &
      2159                           &
      10.00
      \\
      default of credit card clients &
      25700                          &
      23                             &
      23364                          &
      2336                           &
      10.00
      \\
      nomao                          &
      25852                          &
      119                            &
      24621                          &
      1231                           &
      20.00
      \\
      \bottomrule
    \end{tabular}%
  }
\end{table}%

\begin{table}[htbp]
  \centering
  \caption{Information of the multi-class datasets}
  \label{datasets_multi}
  \scalebox{0.79}{
    \begin{tabular}{lcccc}
    \toprule
    Datasets & Samples & Features & Classes & Class distribution(descending) \\
    \midrule
    wine  & 178   & 13    & 3     & 71: 59: 48 \\
    vertebralColumn & 310   & 6     & 3     & 150: 100: 60 \\
    OBS   & 1075  & 21    & 4     & 500: 300: 155: 120 \\
    PhishingData & 1353  & 9     & 3     & 702: 548: 103 \\
    abalone & 4177  & 8     & 3     & 1528: 1342: 1307 \\
    sensorReadings & 5456  & 24    & 4     & 2205: 2097: 826: 328 \\
    frogs & 7195  & 21    & 4     & 4420: 2165: 542: 68 \\
    \bottomrule
    \end{tabular}%
  }
\end{table}%

\subsubsection{Reference Resampling Algorithms and Frameworks.} To verify the effectiveness and superiority of the proposed INGB framework, extensive comparative experiments are carried out with ten mainstream sampling algorithms, including SMOTE (S.)\cite{bib16}, SMOTE-ENN (S-ENN), SMOTE-TomekLinks (S-TkL)\cite{bib29}, MWMOTE\cite{bib27}, SMOTE-IPF (S-IPF)\cite{bib30}, kmeans-SMOTE(kmeans-S)\cite{bib25}, RSMOTE\cite{bib33}, Adaptive-SMOTE (Adapt-S)\cite{bib28}, Gaussian distribution-based oversampling (GDO)\cite{bib2}, granular-ball sampling (GBS)\cite{bib31} and three state-of-the-art frameworks G-SMOTE \cite{bib36}, W-SMOTEs\cite{bib34}, and SW \cite{bib35}. The implementations of these reference algorithms and frameworks are provided from the SMOTE-variants library\cite{bib41} and the corresponding authors.

\subsubsection{Implementation and Reproducibility.} The experiment was implemented in the Python programming language on an Intel i7-10700 CPU and 32 GB RAM with Windows 10. The complete code required to reproduce the experiment has been made public at \href{https://github.com/dream-lm/INGB\_framework}{https://github.com/dream-lm/INGB\_framework}. Additionally, we provide the average results for each dataset, which can be utilized for further analysis.

\subsection{Experimental Evaluation}
\subsubsection{Classification} To avoid being limited to a specific classifier, six classifiers representing distinct learning paradigms are used throughout the experiments: the logistic regression (LR), $k$-nearest neighbor (KNN), decision tree (DTree), adaptive boosting (AdaBoost), gradient boosting decision tree (GBDT) and multi-layer perceptron (MLP) classifiers. Implementations of all classifiers with default hyperparameters are available from the scikit-learn machine learning library\cite{bib42}.
\subsubsection{Evaluation Metrics}Ten-fold stratified cross-validation was performed on each dataset. Additionally, traditional evaluation metrics such as accuracy that reflect the overall classification performance are no longer applicable. Due to class imbalance, seemingly high accuracy may be achieved without correctly identifying the minority class \cite{bibmetrics}. Therefore, three more comprehensive and suitable evaluation metrics \cite{bib43}, that is, the area under receiving operator characteristics curve (AUC) and the G-mean are listed in Eqs (\ref{eq_pre})-(\ref{eq_gmean}).

\begin{equation}\label{eq_pre}
  \text { Precision }=\frac{TP}{TP+FP},
\end{equation}
\begin{equation}\label{eq_rec}
  \text { Recall/Sensitivity }=\frac{TP}{TP+FN},
\end{equation}
\begin{equation}\label{eq_f1}
  \mathrm{F_{\beta}}\!-\!\!\text { measure }=\frac{(1 + \beta) * \text {Recall} * \text {Precision}}{\beta^{2} * \text {Recall}+\text {Precision}},
\end{equation}
\begin{equation}\label{eq_spec}
  \text {Specificity}=\frac{TN}{TN+FP},
\end{equation}
\begin{equation}\label{eq_auc}
  \mathrm{ AUC }=\frac{\text{Sensitivity} + \text{Specificity}}{2}.
\end{equation}
\begin{equation}\label{eq_gmean}
  \mathrm{G}\!-\!\!\text { mean }=\sqrt{\text{Sensitivity} * \text{Specificity}}.
\end{equation}

Among them, TP (TN) and FP (FN) represent the number of correctly and mistakenly classified minority (majority) classes, respectively. Precision represents the proportion of the true minority class among all samples predicted to be the minority, and reflects the accurate prediction of the minority class. Recall represents the proportion of the true minority samples that is correctly predicted out of all minority samples, and reflects the correct search of the minority. Specificity is the same as recall, which reflects the correct search of the majority classes. Therefore, the F1-measure is a trade-off between precision and recall for the minority class, and AUC and G-mean emphasize the recall of multiple classes.

\begin{figure*}
  \centering
  \includegraphics[scale=0.48]{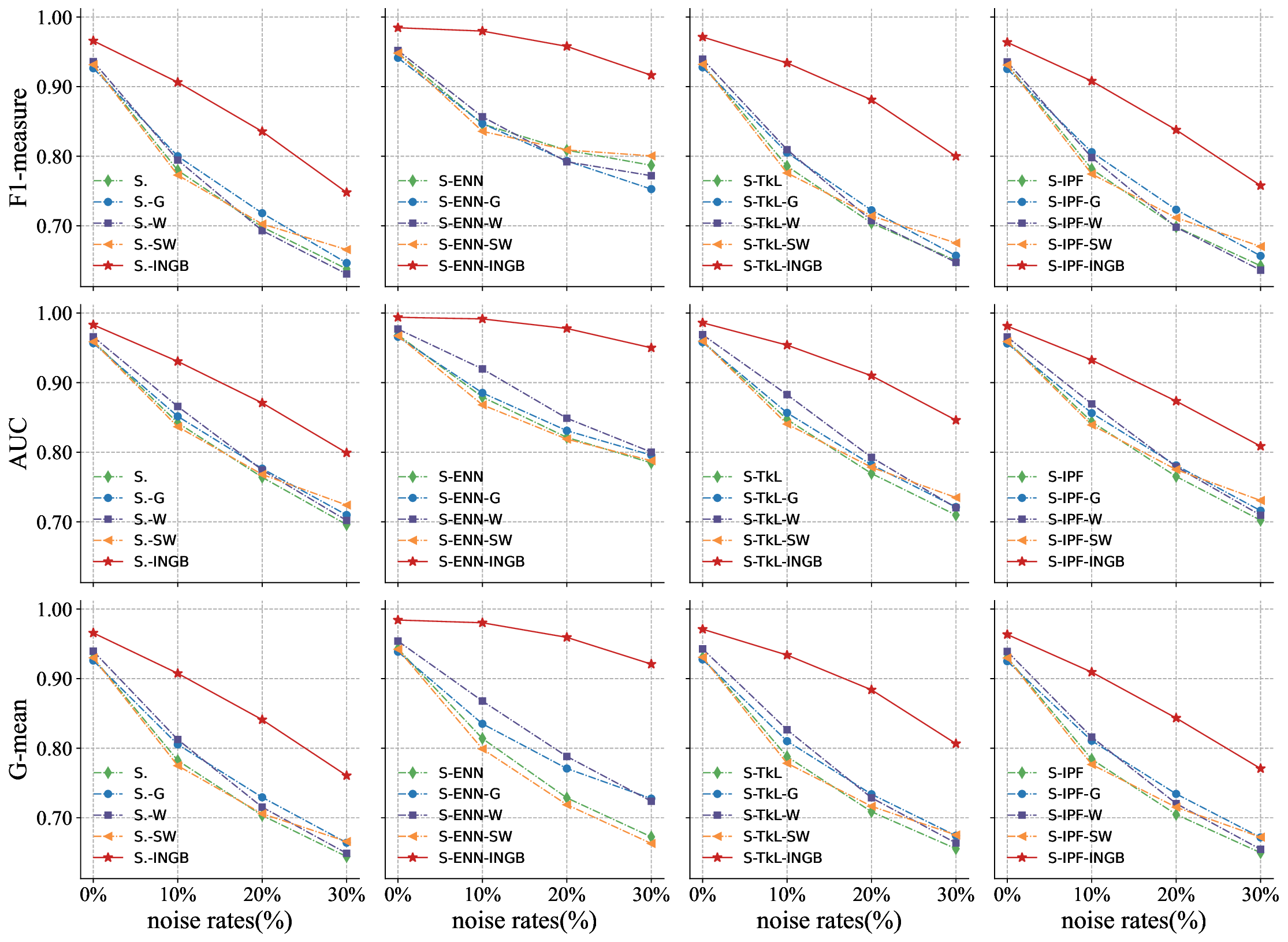}
  \caption{The average results of six classifiers on 26 real-world datasets for framework comparison with varying noise rates (0\% $\leq \ nr \leq $ 30\%) }
  \label{framework}
\end{figure*}

\subsection{Comparison with State-of-the-art Frameworks for Improving Representative Algorithms} \label{experiment_framework}
To demonstrate the effectiveness and robustness of the INGB framework, three oversampling frameworks including G-SMOTE, W-SMOTEs and SW, are compared on four representative algorithms: S., S-ENN, S-TkL and S-IPF. The average results of six classifiers on 22 real-world binary datasets with varying noise rates (0\% $\leq \ nr \leq $ 30\%) are shown in Fig. \ref{framework}. "G.", "W." and "SW." denote the abbreviation of the corresponding framework combined with other algorithms.
\subsubsection{Analysis of Representative Algorithms Improvements}
The diamond-shaped green dotted line and the five-star red marker in each subfigure of Fig. \ref{framework} represent the original representative algorithm and its INGB-improved algorithm, respectively. Overall, the solid red line is consistently above the dashed green line for all metrics, algorithms and noise rates. This demonstrates that the INGB framework is capable of enhancing representative algorithms irrespective of the measurements, methodologies and noise rates.
It is well known that the presence of noise increases the complexity of learning and reduces the prediction performance \cite{bibnoise}, so the performance of the algorithm decreases with increasing of noise. For each algorithm, the improvement without noise is less than 5\%. However, the original algorithm is significantly enhanced by more than 10\% with even slight noise, especially the improvement of S-ENN-INGB is greater than 20\% under G-mean with noise rates reaching 20\%. The main reasons why INGB can stably enhance the original representative algorithm with noise are as follows: First, the adaptive granular generation strategy continuously monitors the noise information while maintaining high-quality granular partitioning. A complete and efficient data space search can effectively separate safe and noisy regions. Moreover, the density-based informed entropy can not only effectively measure the information content of granular balls, but also filter out noisy low-quality granular balls. Finally, sampling involving $k$NN is inherently more susceptible to class noise.

\begin{table*}[htbp]
  \centering
  \caption{the mean and standard deviation results averaged on six classifiers and 22 binary datasets for algorithm comparison}
  \resizebox{\textwidth}{!}{
    \begin{tabular}{clcccccccccc}
    \toprule
    \multicolumn{1}{l}{noise rates} & Metrics & S.    & MWMOTE & Adapt-S & S-TkL & S-IPF & RSMOTE & kmeans-S & GDO   & GBS   & S.-INGB \\
    \midrule
    \multirow{3}[1]{*}{0\%} & F1-measure & 0.932 ± 0.01 & 0.937 ± 0.01 & 0.943 ± 0.01 & 0.934 ± 0.01 & 0.933 ± 0.01 & 0.955 ± 0.01 & 0.960 ± 0.01 & 0.947 ± 0.01 & 0.654 ± 0.06 & \textbf{0.966 ± 0.01} \\
          & AUC   & 0.960 ± 0.01 & 0.964 ± 0.01 & 0.967 ± 0.01 & 0.962 ± 0.01 & 0.961 ± 0.01 & 0.979 ± 0.01 & 0.979 ± 0.01 & 0.972 ± 0.01 & 0.885 ± 0.03 & \textbf{0.983 ± 0.00} \\
          & G-mean & 0.930 ± 0.01 & 0.936 ± 0.01 & 0.939 ± 0.01 & 0.932 ± 0.01 & 0.931 ± 0.01 & 0.956 ± 0.01 & 0.960 ± 0.01 & 0.947 ± 0.01 & 0.706 ± 0.06 & \textbf{0.966 ± 0.01} \\
    \multirow{3}[0]{*}{10\%} & F1-measure & 0.780 ± 0.02 & 0.793 ± 0.02 & 0.844 ± 0.02 & 0.785 ± 0.02 & 0.780 ± 0.02 & 0.881 ± 0.02 & 0.863 ± 0.02 & 0.869 ± 0.02 & 0.420 ± 0.07 & \textbf{0.906 ± 0.02} \\
          & AUC   & 0.842 ± 0.02 & 0.848 ± 0.02 & 0.873 ± 0.02 & 0.846 ± 0.02 & 0.844 ± 0.02 & 0.919 ± 0.01 & 0.905 ± 0.02 & 0.903 ± 0.02 & 0.716 ± 0.05 & \textbf{0.930 ± 0.01} \\
          & G-mean & 0.782 ± 0.02 & 0.795 ± 0.02 & 0.825 ± 0.02 & 0.787 ± 0.02 & 0.783 ± 0.02 & 0.885 ± 0.02 & 0.864 ± 0.02 & 0.871 ± 0.02 & 0.518 ± 0.07 & \textbf{0.908 ± 0.02} \\
    \multirow{3}[0]{*}{20\%} & F1-measure & 0.698 ± 0.03 & 0.713 ± 0.03 & 0.811 ± 0.02 & 0.705 ± 0.03 & 0.700 ± 0.03 & 0.785 ± 0.03 & 0.790 ± 0.02 & 0.777 ± 0.02 & 0.326 ± 0.06 & \textbf{0.836 ± 0.02} \\
          & AUC   & 0.764 ± 0.02 & 0.768 ± 0.03 & 0.816 ± 0.02 & 0.769 ± 0.02 & 0.766 ± 0.02 & 0.843 ± 0.02 & 0.838 ± 0.02 & 0.824 ± 0.02 & 0.647 ± 0.05 & \textbf{0.871 ± 0.02} \\
          & G-mean & 0.703 ± 0.03 & 0.718 ± 0.03 & 0.751 ± 0.02 & 0.710 ± 0.03 & 0.706 ± 0.02 & 0.795 ± 0.02 & 0.792 ± 0.02 & 0.784 ± 0.02 & 0.440 ± 0.06 & \textbf{0.841 ± 0.02} \\
    \multirow{3}[1]{*}{30\%} & F1-measure & 0.638 ± 0.03 & 0.653 ± 0.03 & \textbf{0.802 ± 0.02} & 0.649 ± 0.03 & 0.643 ± 0.03 & 0.668 ± 0.03 & 0.720 ± 0.03 & 0.692 ± 0.03 & 0.271 ± 0.06 & 0.748 ± 0.03 \\
          & AUC   & 0.696 ± 0.03 & 0.708 ± 0.03 & 0.774 ± 0.02 & 0.708 ± 0.03 & 0.702 ± 0.03 & 0.749 ± 0.03 & 0.775 ± 0.03 & 0.750 ± 0.03 & 0.591 ± 0.04 & \textbf{0.799 ± 0.02} \\
          & G-mean & 0.644 ± 0.03 & 0.661 ± 0.03 & 0.656 ± 0.03 & 0.655 ± 0.03 & 0.651 ± 0.03 & 0.689 ± 0.03 & 0.729 ± 0.03 & 0.707 ± 0.03 & 0.390 ± 0.06 & \textbf{0.761 ± 0.02} \\
    \bottomrule
    \end{tabular}%
  \label{algorithm_comparison}%
  }
\end{table*}%

\subsubsection{Analysis of Comparing with State-of-the-art Frameworks}
The red solid and dashed lines in each subplot of Fig. \ref{framework} represent the algorithms improved by INGB and the other frameworks, respectively. Overall, for all metrics, algorithms, and noise rates, all dashed lines remain near the solid green line but are consistently well below the solid red line. This reveals that the INGB framework can steadily outperform the state-of-the-art frameworks regardless of the metrics, algorithms and noise rates. Specifically, compared to the other metrics, all frameworks achieve the highest scores for AUC, but the improvement in AUC is not that great. As illustrated by the extensive experiments on the imbalanced dataset in \cite{bibexperiments}, it is considerably more difficult to enhance the results of rank-based AUC compared to TPR and TNR-based criteria. However, the INGB framework remains steadily improving. Furthermore, compared to the other algorithms, all frameworks achieve the greatest scores in SMOTE-ENN. The compared frameworks can only marginally improve or lag slightly, however the INGB framework advances the most.
The main reasons why the INGB is comprehensively superior to the other frameworks are as follows: First, adaptive granular generation and density-based information entropy effectively monitor and reduce the damage caused by noise. Second, the number of synthetic samples is distributed in accordance with the sparsity of the high-dimensional ball space, which not only alleviates sample overlap but also facilitates the reasonable generalization of the high-dimensional sphere space. Finally, nonlinear granular sampling with a spherical Gaussian distribution not only expands the neighborhood sampling space but also follows the data distribution, generating more informative and diversified new samples. Therefore, INGB can further improve the performance even if it is already satisfactory, as shown by the AUC metric and the S-ENN algorithm.

\subsection{Comparison with Mainstream Sampling Algorithms}
Since the INGB framework already outperforms other frameworks under diverse representative sampling methods, S.-INGB with lower performance improvement but simplicity is chosen for comparison with other mainstream sampling methods. Among them, S., MWMOTE and Adapt-S are rea-emphasizing sampling algorithms. S-TkL, S-IPF and RSMOTE are noise-filtering algorithms. kmeans-S and RSMOTE are cluster-based sampling algorithms. GDO is a distribution-based sampling algorithm. GBS is a robust granular-based undersampling algorithm. The average algorithm comparison results of six classifiers on 22 real-world binary datasets with varying noise rates (0\% $\leq \ nr \leq $ 30\%) are shown Table \ref{algorithm_comparison}.

From Table \ref{algorithm_comparison}, it is noted that S.-INGB surpasses all other sampling algorithms for all noise rates and metrics, with the exception of the F1-measure of Adapt-S with the 30\% noise rate. More specifically, the F1-measure of Adapt-S is 5.4\% higher than that of S.-INGB with a noise rate of 30\%, but the AUC and G-mean scores of Adapt-S are lower than those of S.-INGB. for example, the G-mean score is 10.5\% lower. One possible reason is that Adapt-S has meticulously created distinct synthesis strategies for noise, whereas the S.-INGB does not impose any restrictions to allow synthesized samples to be more similar to real data distribution and preserve sample diversity. Hence, it is not as effective as Adapt-S at the 30 \% noise rate. Another crucial reason is that the AUC and G-mean metrics emphasize the overall recall of multiple classes, but the F1-measure focuses on the precision and recall of minority classes only \cite{bibmetrics2}. This indicates that the improvement gained by Adapt-S in emphasizing the minority class may degrade the majority class results. In contrast, S.-INGB generates more diversified new samples within a reasonable minority class distribution. Moreover, the average results of GBS are not satisfactory. The primary issue may be that undersampling is likely to cause the loss of valuable information while discarding majority samples, and the effect of GBS is dependent on the adjustment of purity hyperparameters, which may be difficult to guarantee for some complicated datasets. Except for Adapt-S with 10\% noise, RSMOTE and kmeans-S with 20\% noise achieve the maximum improvement. S.-INGB achieves the continuously increasing improvements as the noise rate increases. This phenomenon is consistent with that described in subsection \ref{experiment_framework}. These results reveal that not only does the INGB framework outperform mainstream algorithms, but it is also suitable for high-noise imbalanced classification.

\begin{table}[htbp]
  \centering
  \caption{Adjusted p-values of friedman's test}
  \scalebox{0.75}{
    \begin{tabular}{cccccccc}
      \toprule
      Experiments                                 &
      \multicolumn{3}{c}{Framework comparisons}   &
                                                  &
      \multicolumn{3}{c}{Algorithm comparisons}
      \\
      \cmidrule{2-4}\cmidrule{6-8}    Noise rates &
      F1-measure                                  &
      AUC                                         &
      G-mean                                      &
                                                  &
      F1-measure                                  &
      AUC                                         &
      G-mean
      \\
      \midrule
      0\%                                         &
      2.00E-07 &	6.29E-07 &	1.81E-07 &
                                                  &
      1.92E-11	& 5.29E-09	& 5.33E-10
      \\
      10\%                                        &
      1.24E-18 &	1.77E-18 &	6.04E-19 &

                                                  &
      2.99E-39 &	1.94E-29	& 7.66E-37

      \\
      20\%                                        &
      2.49E-22 &	6.83E-19 &	9.50E-20 &
                                                  &
      4.90E-45	& 4.77E-36	& 2.70E-36

      \\
      30\%                                        &
      1.34E-24 &	1.80E-19 &	1.77E-18 &
                                                  &
      3.90E-50 &	4.46E-44	& 1.47E-36

      \\
      \bottomrule
    \end{tabular}%
    \label{friedman}%
  }
\end{table}%

\begin{figure}
  \centering
  \includegraphics[scale=0.29]{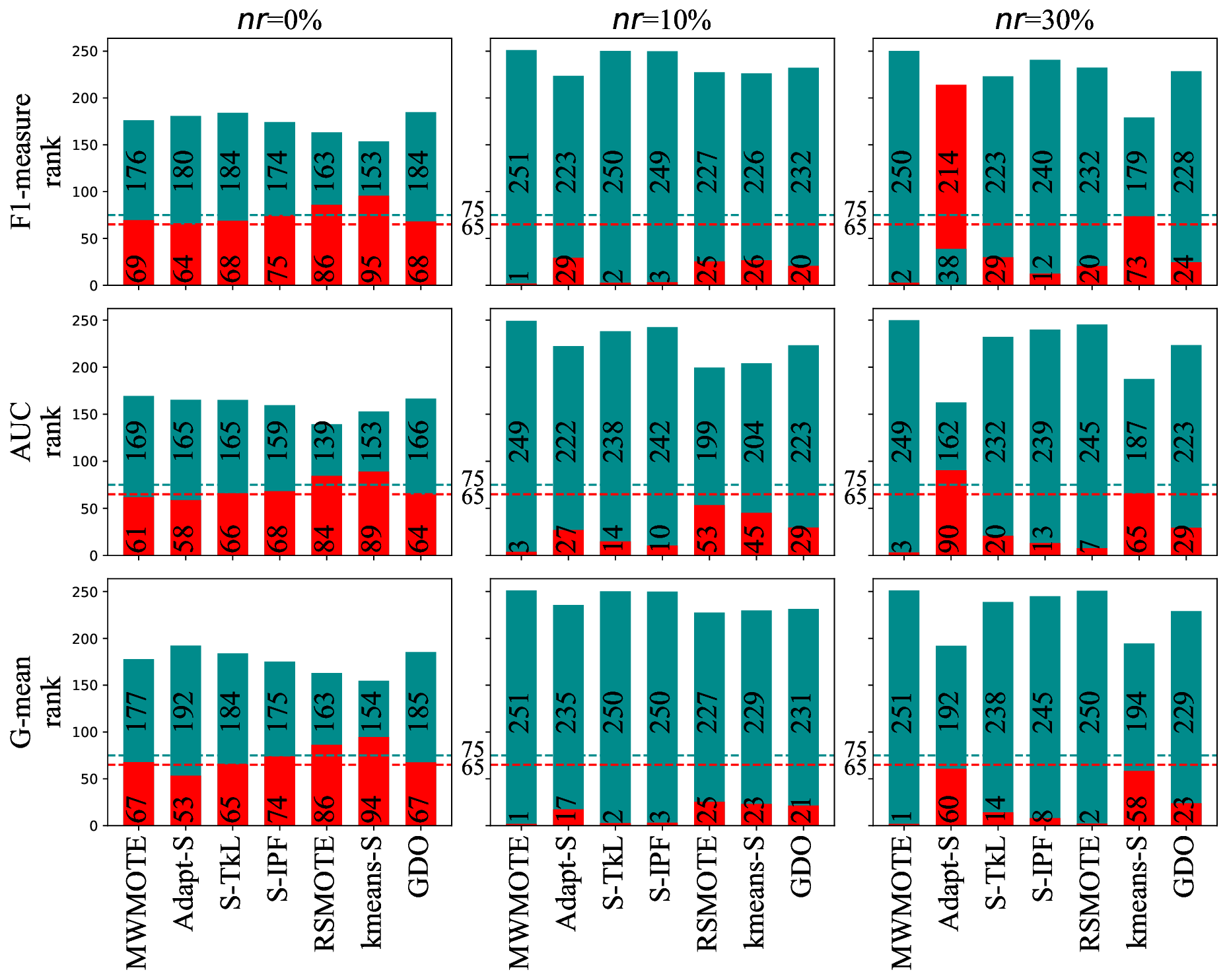}
  \caption{ The average visual results of wilcoxon signed-rank test for algorithm comparisons}
  \label{wilcoxon_vsother}
\end{figure}

\begin{figure*}
  \centering
  \includegraphics[scale=0.42]{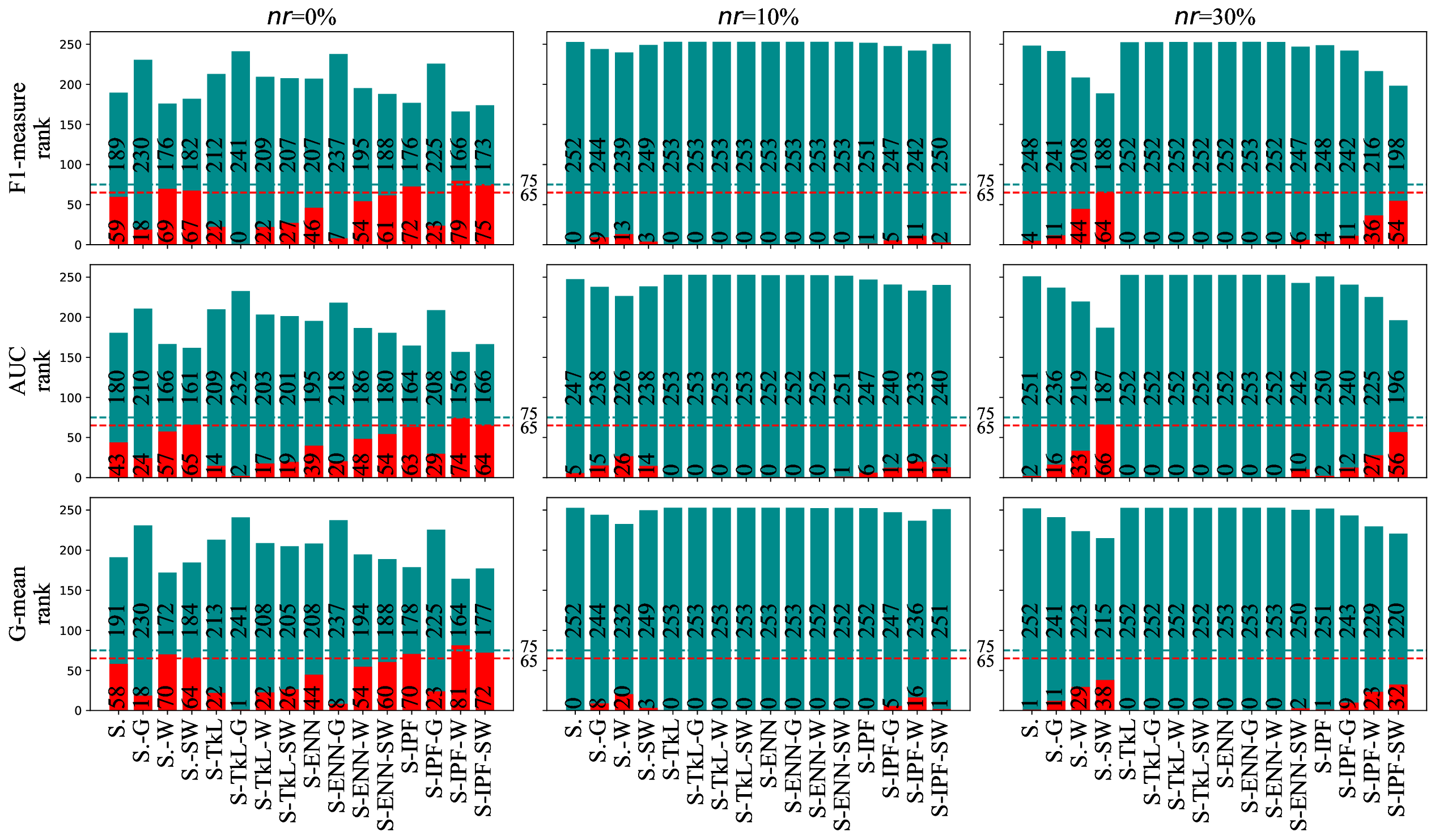}
  \caption{The average visual results of wilcoxon signed-rank test for framework comparisons }
  \label{wilcoxon_framework}
\end{figure*}

\begin{table*}[htbp]
  \centering
  \caption{the mean and standard deviation results averaged on six classifiers and seven multi-class datasets for framework comparison}
  \scalebox{0.95}{
    \begin{tabular}{clcccccccc}
    \toprule
    \multicolumn{1}{l}{Noise rates} & Metrics & S. & S.-INGB & S-ENN & S-ENN-INGB & S-TkL & S-TkL-INGB & S-IPF & S-IPF-INGB \\
    \midrule
    \multirow{3}[1]{*}{0\%} & F1-measure & 0.789 ± 0.04 & \textbf{0.835 ± 0.03} & 0.839 ± 0.02 & \textbf{0.888 ± 0.03} & 0.796 ± 0.03 & \textbf{0.836 ± 0.03} & 0.909 ± 0.02 & \textbf{0.920 ± 0.02} \\
          & AUC   & 0.913 ± 0.02 & \textbf{0.930 ± 0.01} & 0.941 ± 0.01 & \textbf{0.956 ± 0.01} & 0.917 ± 0.02 & \textbf{0.936 ± 0.01} & 0.928 ± 0.01 & \textbf{0.936 ± 0.01} \\
          & G-mean & 0.793 ± 0.04 & \textbf{0.838 ± 0.03} & 0.844 ± 0.02 & \textbf{0.892 ± 0.03} & 0.800 ± 0.03 & \textbf{0.839 ± 0.03} & 0.908 ± 0.02 & \textbf{0.919 ± 0.02} \\
    \multirow{3}[0]{*}{10\%} & F1-measure & 0.717 ± 0.04 & \textbf{0.799 ± 0.03} & 0.822 ± 0.04 & \textbf{0.882 ± 0.03} & 0.735 ± 0.04 & \textbf{0.829 ± 0.03} & 0.856 ± 0.03 & \textbf{0.893 ± 0.03} \\
          & AUC   & 0.852 ± 0.03 & \textbf{0.896 ± 0.02} & 0.935 ± 0.02 & \textbf{0.957 ± 0.01} & 0.869 ± 0.02 & \textbf{0.917 ± 0.02} & 0.897 ± 0.02 & \textbf{0.913 ± 0.02} \\
          & G-mean & 0.722 ± 0.04 & \textbf{0.801 ± 0.03} & 0.827 ± 0.04 & \textbf{0.885 ± 0.03} & 0.741 ± 0.04 & \textbf{0.831 ± 0.03} & 0.857 ± 0.03 & \textbf{0.888 ± 0.03} \\
    \multirow{3}[0]{*}{20\%} & F1-measure & 0.629 ± 0.04 & \textbf{0.744 ± 0.03} & 0.786 ± 0.05 & \textbf{0.884 ± 0.03} & 0.654 ± 0.04 & \textbf{0.778 ± 0.04} & 0.808 ± 0.04 & \textbf{0.864 ± 0.03} \\
          & AUC   & 0.789 ± 0.03 & \textbf{0.859 ± 0.02} & 0.920 ± 0.02 & \textbf{0.953 ± 0.01} & 0.812 ± 0.03 & \textbf{0.878 ± 0.03} & 0.858 ± 0.03 & \textbf{0.884 ± 0.03} \\
          & G-mean & 0.636 ± 0.04 & \textbf{0.748 ± 0.03} & 0.792 ± 0.05 & \textbf{0.887 ± 0.03} & 0.661 ± 0.04 & \textbf{0.781 ± 0.04} & 0.809 ± 0.04 & \textbf{0.860 ± 0.03} \\
    \multirow{3}[1]{*}{30\%} & F1-measure & 0.573 ± 0.04 & \textbf{0.690 ± 0.04} & 0.763 ± 0.05 & \textbf{0.858 ± 0.04} & 0.599 ± 0.04 & \textbf{0.742 ± 0.04} & 0.778 ± 0.04 & \textbf{0.831 ± 0.04} \\
          & AUC   & 0.748 ± 0.03 & \textbf{0.820 ± 0.02} & 0.903 ± 0.03 & \textbf{0.939 ± 0.02} & 0.772 ± 0.03 & \textbf{0.855 ± 0.03} & 0.824 ± 0.04 & \textbf{0.860 ± 0.03} \\
          & G-mean & 0.581 ± 0.04 & \textbf{0.695 ± 0.04} & 0.769 ± 0.05 & \textbf{0.860 ± 0.04} & 0.606 ± 0.04 & \textbf{0.746 ± 0.04} & 0.777 ± 0.04 & \textbf{0.825 ± 0.04} \\
    \bottomrule
    \end{tabular}%
  \label{framework_multi}%
  }
\end{table*}%

\begin{table*}[htbp]
  \centering
  \caption{the mean and standard deviation results averaged on six classifiers and seven multi-class datasets for algorithm comparison}
  \resizebox{\textwidth}{!}{
    \begin{tabular}{clccccccc}
    \toprule
    \multicolumn{1}{l}{Noise rates} & Metrics & S.    & MWMOTE & S-TkL & RSMOTE & kmeans-S & GBSY  & S.-INGB \\
    \midrule
    \multirow{3}[1]{*}{0\%} & F1-measure & 0.789 ± 0.04 & 0.799 ± 0.03 & 0.793 ± 0.03 & 0.779 ± 0.02 & 0.820 ± 0.03 & 0.643 ± 0.04 & \textbf{0.835 ± 0.03} \\
          & AUC   & 0.913 ± 0.02 & 0.917 ± 0.01 & 0.915 ± 0.01 & 0.911 ± 0.01 & 0.926 ± 0.01 & 0.831 ± 0.03 & \textbf{0.930 ± 0.01} \\
          & G-mean & 0.793 ± 0.04 & 0.802 ± 0.03 & 0.797 ± 0.02 & 0.786 ± 0.02 & 0.823 ± 0.02 & 0.648 ± 0.04 & \textbf{0.838 ± 0.03} \\
    \multirow{3}[0]{*}{10\%} & F1-measure & 0.717 ± 0.04 & 0.719 ± 0.04 & 0.731 ± 0.04 & 0.751 ± 0.04 & 0.788 ± 0.03 & 0.513 ± 0.05 & \textbf{0.799 ± 0.03} \\
          & AUC   & 0.852 ± 0.03 & 0.852 ± 0.03 & 0.866 ± 0.02 & 0.879 ± 0.02 & \textbf{0.896 ± 0.02} & 0.713 ± 0.05 & \textbf{0.896 ± 0.02} \\
          & G-mean & 0.722 ± 0.04 & 0.724 ± 0.04 & 0.736 ± 0.04 & 0.757 ± 0.03 & 0.790 ± 0.03 & 0.520 ± 0.05 & \textbf{0.801 ± 0.03} \\
    \multirow{3}[0]{*}{20\%} & F1-measure & 0.629 ± 0.04 & 0.640 ± 0.04 & 0.653 ± 0.04 & 0.688 ± 0.04 & 0.708 ± 0.04 & 0.468 ± 0.05 & \textbf{0.744 ± 0.04} \\
          & AUC   & 0.789 ± 0.03 & 0.795 ± 0.03 & 0.810 ± 0.03 & 0.833 ± 0.02 & 0.844 ± 0.02 & 0.678 ± 0.05 & \textbf{0.859 ± 0.02} \\
          & G-mean & 0.636 ± 0.04 & 0.645 ± 0.04 & 0.660 ± 0.04 & 0.693 ± 0.04 & 0.711 ± 0.04 & 0.477 ± 0.05 & \textbf{0.748 ± 0.04} \\
    \multirow{3}[1]{*}{30\%} & F1-measure & 0.573 ± 0.04 & 0.582 ± 0.05 & 0.595 ± 0.04 & 0.634 ± 0.04 & 0.645 ± 0.04 & 0.438 ± 0.05 & \textbf{0.690 ± 0.04} \\
          & AUC   & 0.748 ± 0.03 & 0.749 ± 0.03 & 0.769 ± 0.03 & 0.792 ± 0.03 & 0.798 ± 0.03 & 0.651 ± 0.05 & \textbf{0.820 ± 0.03} \\
          & G-mean & 0.581 ± 0.04 & 0.588 ± 0.05 & 0.602 ± 0.04 & 0.640 ± 0.04 & 0.648 ± 0.04 & 0.448 ± 0.05 & \textbf{0.695 ± 0.04} \\
    \bottomrule
    \end{tabular}%
  \label{methods_multi}%
  }
\end{table*}%

\subsection{Nonparametric Tests}
To demonstrate the statistical effectiveness of the INGB framework, rank-based nonparametric tests, Friedman's test and Wilcoxon signed-ranks test are used to test hypotheses about multivariate relationships and differences \cite{bibstact1}. Friedman's test is used within the framework and algorithm comparisons to detect whether multiple algorithms were significantly different. If the probability p-value is less than the given significance level, the null hypothesis is rejected, and the algorithms are considered to be significantly different. The corresponding adjusted p-values are given in Table \ref{friedman}. Overall, INGB achieves significant improvements over other frameworks and algorithms in noise-imbalanced learning, as the adjusted p-values for the tests are all less than 0.005.

Furthermore, the Wilcoxon signed-rank test is used to precisely evaluate the performance difference for each algorithm and framework pair \cite{bibstact2}. The average visual results of the Wilcoxon signed-rank test for the four groups of framework comparisons and the nine pairs of mainstream algorithms on six classifiers and 22 datasets are depicted in Figs. \ref{wilcoxon_framework} and \ref{wilcoxon_vsother}, respectively, where blue bars denote the positive rank sums for all datasets in which the INGB framework beats the paired algorithms or frameworks and red bars represent the inverse. Since 22 datasets were utilized, the critical values for significance level $\alpha = 0.05$ and $\alpha = 0.1$ should not exceed 65 (marked by the red line) and 75 (marked by the blue line), respectively. From Fig \ref{wilcoxon_framework}, the INGB framework can clearly reject the null hypothesis at a significance level = 0.05, except for S-IPF-W when noise is 0 at a significance level = 0.1. From Fig \ref{wilcoxon_vsother}, S.-INGB can clearly reject the null hypothesis at a significance level = 0.1, except for RSMOTE and kmeans-S with noise rate of 0\%. Adapt-S cannot reject the null hypothesis in terms of the F1-measure and AUC with 30\% noise. In conclusion, the results of the statistical tests demonstrate that the INGB framework significantly surpasses the state-of-the-art frameworks and mainstream algorithms.

\subsection{Experiments on Imbalanced Multi-Class DataSets}
To further study the extensibility of the INGB framework on multi-class classification problems, six mainstream algorithms are utilized for algorithm comparison on seven multi-class datasets. The details of the multi-class dataset are provided in Table \ref{datasets_multi}. Since the three frameworks mentioned above are not suitable for multi-classification problems, four representative algorithms are utilized for the framework comparison. Tables \ref{framework_multi}-\ref{methods_multi} show the average results of six classifiers on seven real-world multi-class datasets with varying noise rates (0\% $\leq \ nr \leq $ 30\%). From Table \ref{framework_multi}, the INGB framework not only steadily outperforms comparative representative algorithms regardless of the noise rates, metrics, and algorithms, but also maintains a higher improvement under noise. From Table \ref{methods_multi}, S.-INGB not only achieves the best performance regardless of the noise rates and metrics but also maintains more significant advancements under noise, except for the AUC scores of kmeans-S with 10\% noise. The extensive experimental results demonstrate that the INGB framework maintains superiority on multi-classification problems. Therefore the proposed INGB framework is suitable for handling imbalanced binary classification and imbalanced multi-classification problems, especially high-noise imbalance.

\section{Conclusion} \label{conclusion}
In this article, an informed nonlinear oversampling framework and related mathematical models are proposed for noisy imbalance classification. First, adaptive granular-ball generation obtains comprehensive and high-quality balls that characterize the data distribution, thereby not only mitigating the negative effects of noisy samples and $k$-nearest neighbor hyperparameter tuning, but also extending the instance-level problem to nonlinear granular spaces. Second, density-based informed entropy can further optimize the granular space's information richness and robustness. Finally, diversified and well-informed nonlinear sampling follows sparse and spherical Gaussian distributions. To verify the effectiveness and robustness of the proposed INGB framework, extensive experiments are implemented on six classifiers and 31 real-world datasets with varying noise rates, including the comparison of state-of-the-art frameworks on representative algorithms, the comparison of mainstream algorithms and several statistical analyses. The experimental results demonstrate that the INGB framework significantly outperforms other mainstream frameworks and algorithms, especially in the presence of noise. In the future, the results encourage us to continue working on real-world applications to address complex imbalanced data problems. Furthermore, we shall apply the proposed framework to address high-dimensional noise imbalance problems and explore high-dimensional convex optimizations to further enhance performance.

\section*{Acknowledgments}
This work is supported by the Key Cooperation Project of Chongqing Municipal Education Commission(HZ2021008), partly funded by the State Key Program of National Nature Science Foundation of China (61936001), National Nature Science Foundation of China (61772096), the Key Research and Development Program of Chongqing (cstc2017zdcyzdyfx0091) and the Key Research and Development Program on AI of Chongqing (cstc2017rgznzdyfx0022).

\bibliographystyle{IEEEtran}
\bibliography{IEEEabrv, reference.bib}

\end{document}